\documentclass[conference]{IEEEtran}
\IEEEoverridecommandlockouts
\usepackage{cite}
\usepackage{amsmath,amssymb,amsfonts}
\usepackage{algorithmic}
\usepackage{graphicx}
\usepackage{textcomp}
\usepackage{xcolor}
\usepackage{multicol}
\PassOptionsToPackage{hyphens}{url}\usepackage{hyperref}

\usepackage{tikz}
\newcommand\copyrighttext{
    \footnotesize \textcopyright~2021 IEEE. Personal use of this material is permitted. Permission from IEEE must be obtained for all other uses, in any current or future media, including reprinting/republishing this material for advertising or promotional purposes, creating new collective works, for resale or redistribution to servers or lists, or reuse of any copyrighted component of this work in other works. The final version of this article is available at: https://doi.org/10.1109/ICACSIS53237.2021.9631364
}
\newcommand\copyrightnotice{
    \begin{tikzpicture}[remember picture, overlay]
        \node[anchor=south, yshift=10pt] at (current page.south) {
            \fbox{
                \parbox{\dimexpr1.0\textwidth-\fboxsep-\fboxrule\relax}{\copyrighttext}
            }
        };
    \end{tikzpicture}
}
\def\BibTeX{{\rm B\kern-.05em{\sc i\kern-.025em b}\kern-.08em
    T\kern-.1667em\lower.7ex\hbox{E}\kern-.125emX}}
\begin{document}

\title{
    Investigating Text Shortening Strategy in BERT: Truncation vs Summarization
}

\author{
    \IEEEauthorblockN{Mirza Alim Mutasodirin\IEEEauthorrefmark{1}, Radityo Eko Prasojo\IEEEauthorrefmark{1}\IEEEauthorrefmark{2}}
    \IEEEauthorblockA{\IEEEauthorrefmark{1}Faculty of Computer Science, Universitas Indonesia}
    \IEEEauthorblockA{\IEEEauthorrefmark{2}Kata.ai Research Team, Jakarta, Indonesia}
    \IEEEauthorblockA{\IEEEauthorrefmark{1}mirza.alim01@ui.ac.id, \IEEEauthorrefmark{2}ridho@kata.ai}
}

\maketitle
\copyrightnotice
\begin{abstract}
% Background
The parallelism of Transformer-based models comes at the cost of their input max-length.
% Knowledge Gap
Some studies proposed methods to overcome this limitation, but none of them reported the effectiveness of summarization as an alternative.
% Here we show
In this study, we investigate the performance of document truncation and summarization in text classification tasks. Each of the two was investigated with several variations. This study also investigated how close their performances are to the performance of full-text. We used a dataset of summarization tasks based on Indonesian news articles (IndoSum) to do classification tests.
% Result
This study shows how the summaries outperform the majority of truncation method variations and lose to only one. The best strategy obtained in this study is taking the head of the document. The second is extractive summarization.
% Implication/Contribution
This study explains what happened to the result, leading to further research in order to exploit the potential of document summarization as a shortening alternative. The code and data used in this work are publicly available in \url{https://github.com/mirzaalimm/TruncationVsSummarization}.
\end{abstract}


\section{Introduction}

Transformer-based models~\cite{Vaswani2017Attention} allow a degree of parallelism at the cost of their input length limitation. Typically, for tasks requiring texts longer than the maximum limit, they are truncated at the tail. It is unclear how well this performs, as truncation may cut-off important information within the text. Sun et al. \cite{Sun2019BertTextClassification} experimented with other various truncation strategies, cutting-off different parts of the text. They concluded that cutting off the middle part of the text, taking the head and the tail, works best, but it remains unclear how generalizable this approach is.

In this work, we investigate the effectiveness of text summarization in place of the go-to truncation method. The motivation comes from the fact that there is no general consensus on which parts of a given text are actually important. Nevertheless, an unbiased summarization model is potentially able to extract important parts of the text regardless of their original position, resulting in a highly generalized shortening method.

Specifically, in this study, we compare several text shortening strategies for text classification. We use a DistilBERT \cite{Sanh2019DistilBERT} model, a distilled version of BERT-Base \cite{Devlin2019BERT}, on the IndoSum \cite{Kemal2018IndoSum} dataset, comparing 3 different summarization strategies with 7 different truncation strategies. IndoSum is a dataset for news article classification and summarization in Indonesian. Each article in the dataset is categorized into one of six topics (Entertainment, Inspiration, Sport, Showbiz, Headline, and Tech), and has two ground truth summaries, one extractive and one abstractive. Our experiment shows that applying extractive summarization for shortening strategy outperforms the majority of the truncation strategies (losing to only one) in terms of F1-score on the news topic classification downstream task. This shows that summarization is potentially a feasible and generalizable strategy for text shortening for classification, and can be investigated further with other kinds of documents.

To the best of our knowledge, we are the first to investigate text summarization as a text shortening strategy for a Transformer-based model for any language, and the first to benchmark several shortening strategies for long text classification in \textit{Bahasa Indonesia}. This research is important because in addition to making it easier for readers to understand the documents, the summaries are also expected to be reliable to use in classification tasks. Thus, developers do not need to look for other methods that are only specific to text classification.

The rest of this article is organized as follows. Section \ref{sec:previousWork} describes previous work on document shortening strategies in English and Indonesian. Section \ref{sec:method} provides a detailed explanation of the method and dataset used in this study. Finally, the result of this study will be discussed in Section \ref{sec:resultDiscussion} and will be closed with conclusion and recommendation in Section \ref{sec:conclusion}.

\section{Previous Work} \label{sec:previousWork}

There have been many studies conducted on the task of classifying Indonesian texts. Many of them used classical machine learning on Twitter data, such as a study by Sari et al. \cite{Sari2020SentimentAnalysis} using Na\"ive Bayes and Decision Tree, and Tauhid et al. \cite{Tauhid2020SentimentAnalysis} using SVM, Logistic Regression, Na\"ive Bayes, and Decision Tree. In recent years, LSTM artificial neural networks have also been used for classification tasks, for example a study by Gumilang et al. \cite{Gumilang2019NewsClass} experimenting multilabel classification on Indonesia financial systemic risk online news articles using Bi-LSTM, CNN, and Bi-GRU. Several recent studies have begun to use BERT \cite{Devlin2019BERT} for classification tasks in Indonesian. However, all of these Indonesian BERT-based studies used short texts, for example sentiment analysis by Azhar et al. \cite{Azhar2020SentimentAnalysis}, and some downstream tasks of IndoNLU \cite{Wilie2020IndoNLU} and IndoLEM \cite{Koto2020IndoLEM}. To the best of our knowledge, no specific study investigating shortening methods in Indonesian.

In English, Sun et al. \cite{Sun2019BertTextClassification} conducted a text classification experiment using BERT. To handle text that exceeds 512 tokens, he experimented with 3 simple ways namely (1) taking the first 510 tokens from the text, (2) taking the last 510 tokens, and (3) taking the first 128 tokens and the last 382 tokens from the text. Three other ways he tried were the hierarchical methods, namely dividing the text into $k=L/510$, where $L$ is the number of tokens produced by BERT tokenizer and $k$ is the number of document fractions, and then inputting them into BERT to get a representation of each fraction, then using (1) mean pooling, (2) max pooling, and (3) self-attention to combine all fractions. Of all these methods, the one that performs best is truncation that retrieves the combination of the start and the end of the document.

In another study in English, Pappagari et al. \cite{Pappagari2019HierarchicalTransformers} proposed another hierarchical method. Generally, in hierarchical methods, all parts of the text are taken into account, nothing is discarded, but they are broken into several parts and fed into the model to get a representation of each part, then recombined in several different ways. The problem with this method is that the computation is slower and requires more memory. Wang et al. \cite{Wang2021FeatureSelection} proposed a classical method of feature selection to shorten documents. Ding et al. \cite{Ding2020CogLTX} proposed a method to identify key sentences of a document. His method is not a method intended for summarization tasks and it produces different short representations for different tasks.

\section{Method} \label{sec:method}
This section gives detailed explanation on dataset exploration and setting, shortening variations, preprocessing, and text classification experiment in this study.

\subsection{Data Exploration and Dataset Setting}

In order to test the effectiveness of summarization shortening strategy for text classification, a dataset with human-annotated summaries and topic classification is needed. IndoSum \cite{Kemal2018IndoSum} meets all those requirements. Indonesian actually has another summarization dataset, named Liputan6 \cite{Koto2020Liputan6}. Since it has no topic categorization annotation, it doesn't meet our requirements.

\begin{figure}[htbp]
\centerline{\includegraphics[width=0.5\textwidth]{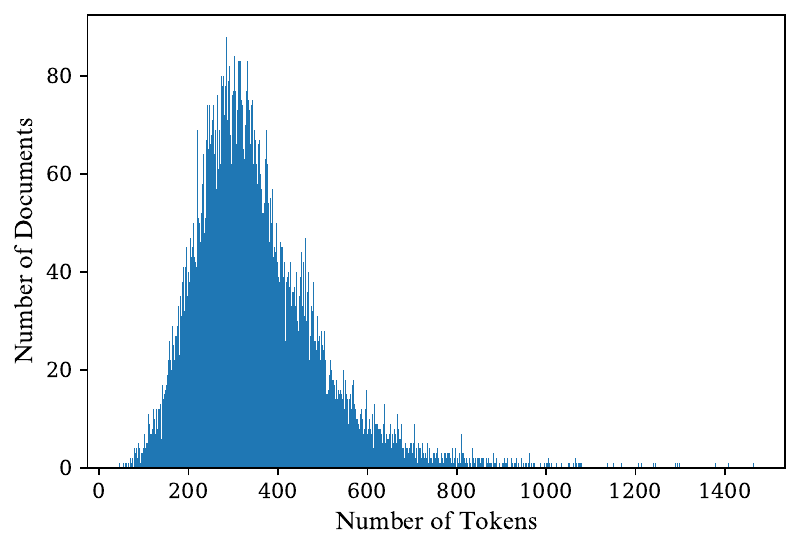}}
\caption{Histogram of IndoSum}
\label{fig:indosumHistogram}
\end{figure}

IndoSum comes with tokenized text. It has a distribution of the number of tokens in a document as shown in Figure \ref{fig:indosumHistogram}. The shortest document has 45 tokens. The longest document has 1464 tokens. The average number of tokens in a document is 346 tokens. IndoSum has a human-annotated extractive summarization with the shortest number of tokens being 8 tokens and the longest being 220 tokens with an average of 75 tokens. While the human-annotated abstractive summarization it has, the shortest article is 37 tokens and the longest is 121 tokens with an average of 69 tokens.

For the purposes of investigating the performance of text classification using the full-text, we filter IndoSum documents and take only those whose length could fit most of the Transformer-based models, which is 512 tokens. Then, we run several text shortening strategies and compare the results, taking the full-text result as the ideal baseline. IndoSum's tokenization could be different from tokenization by the model. Tokenization by the model often breaks words into subwords to deal with out-of-vocabulary cases. Thus, the experiment took only documents with a maximum of 460 tokens, so that if 10\% of the tokens are split into subwords, the full-text can still be fed into the model. If it still exceeds the limit, a bit of the tail can be cut-off.

In order to make the comparison as fair as possible, this experiment set out a text shortening of 70 tokens in length for the truncation method. The number 70 was taken because it is close to the average human-annotated extractive and abstractive summarization that are 75 and 69 tokens respectively. Further, it is preferable that the minimum length of the document is 70x3 tokens (210 tokens), so that the truncation variations of the front, middle, and end of the document do not intersect. Thus, the length of the document taken for the needs of this research is 210 to 460 tokens. We call this dataset Filtered IndoSum.

\begin{figure}[htbp]
\centerline{\includegraphics[width=0.5\textwidth]{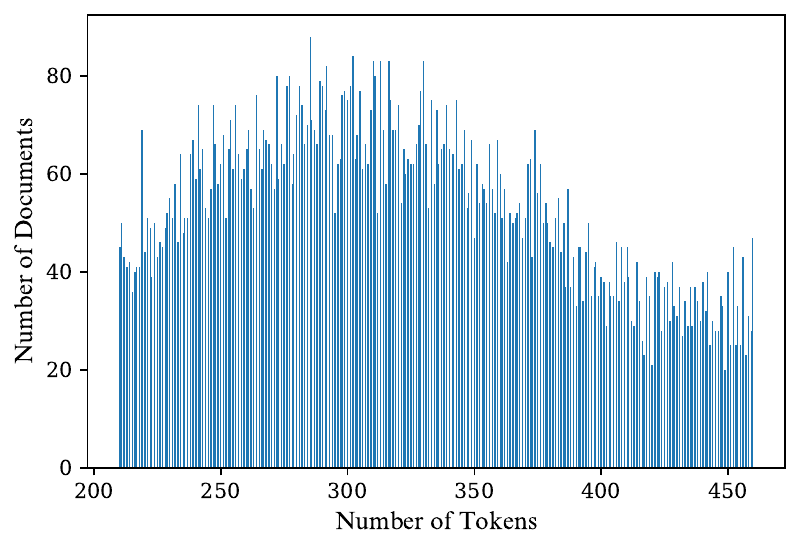}}
\caption{Histogram of Filtered IndoSum}
\label{fig:filteredHistogram}
\end{figure}

\begin{figure}[htbp]
\centerline{\includegraphics[width=0.5\textwidth]{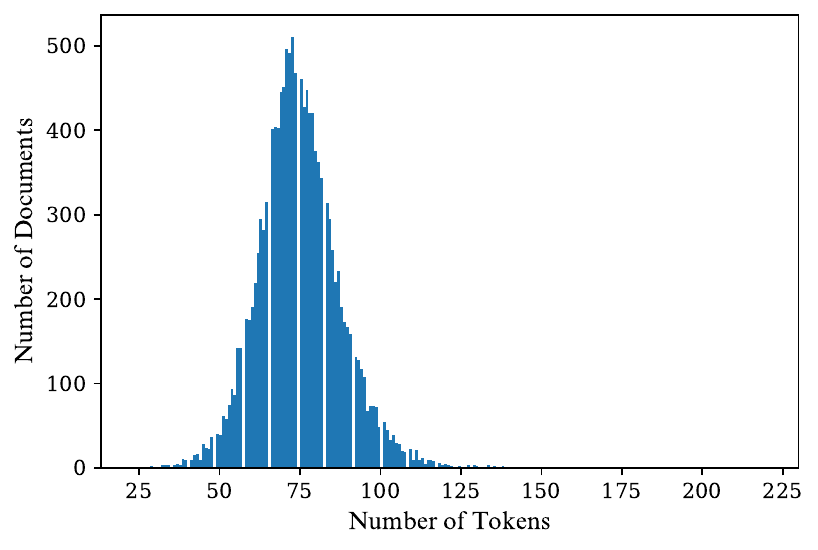}}
\caption{Histogram of Extractive Summarization}
\label{fig:extractiveHistogram}
\end{figure}

\begin{figure}[htbp]
\centerline{\includegraphics[width=0.5\textwidth]{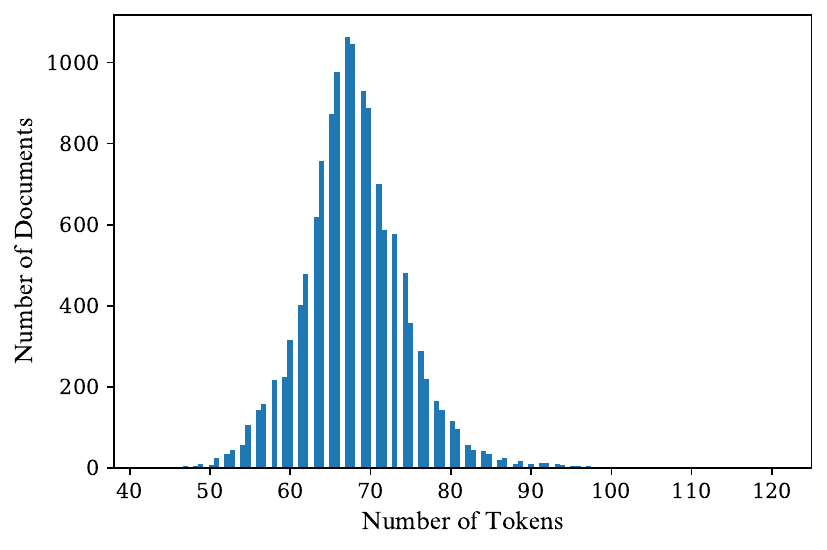}}
\caption{Histogram of Abstractive Summarization}
\label{fig:abstractiveHistogram}
\end{figure}

Filtered IndoSum has a document length distribution as shown in Figure \ref{fig:filteredHistogram}. The total number of documents in Filtered IndoSum is around 13 thousand (13K) articles. The average document length is 323 tokens. The shortest human-annotated extractive summarization is 23 tokens, the longest is 220 tokens, and the average is 75 tokens. Meanwhile, the shortest human-annotated abstractive summarization is 42 tokens, the longest is 121 tokens, and the average is 68 tokens. The distribution is as shown in Figures \ref{fig:extractiveHistogram} and \ref{fig:abstractiveHistogram} respectively.

IndoSum comes in 5-fold cross validation. After the filtering step, we generate a new set of stratified 5 folds, so that each fold has the same label distribution for each train, development, and test set. We also distribute the length of the document in a balanced way so that each of the dataset is not biased towards documents of a certain length. Table \ref{table:crossValidation} shows the data distribution of our new 5 folds.

\begin{table}[htbp]
\caption{Data Distribution of the New 5-Fold Cross Validation}
\begin{center}
\begin{tabular}{lrrrrrrrr}
\hline
 & \multicolumn{6}{c}{\textbf{Category}} & & \\
\textbf{File} & \textbf{1} & \textbf{2} & \textbf{3} & \textbf{4} & \textbf{5} & \textbf{6} & \textbf{Total} & \textbf{\%} \\
\hline
train1 & 4210 & 2717 & 1550 & 1259 & 951 & 60 & \textbf{10747} & 80 \\
dev1 & 261 & 170 & 95 & 77 & 58 & 3 & \textbf{664} & 5 \\
test1 & 794 & 511 & 294 & 240 & 183 & 15 & \textbf{2037} & 15 \\
\textbf{Total} & \textbf{5265} & \textbf{3398} & \textbf{1939} & \textbf{1576} & \textbf{1192} & \textbf{78} & \textbf{13448} & \textbf{100} \\
\hline
train2 & 4210 & 2718 & 1550 & 1259 & 953 & 62 & \textbf{10752} & 80 \\
dev2 & 261 & 169 & 95 & 77 & 58 & 3 & \textbf{663} & 5 \\
test2 & 794 & 511 & 294 & 240 & 181 & 13 & \textbf{2033} & 15 \\
\textbf{Total} & \textbf{5265} & \textbf{3398} & \textbf{1939} & \textbf{1576} & \textbf{1192} & \textbf{78} & \textbf{13448} & \textbf{100} \\
\hline
train3 & 4212 & 2718 & 1550 & 1261 & 953 & 63 & \textbf{10757} & 80 \\
dev3 & 261 & 169 & 95 & 76 & 58 & 2 & \textbf{661} & 5 \\
test3 & 792 & 511 & 294 & 239 & 181 & 13 & \textbf{2030} & 15 \\
\textbf{Total} & \textbf{5265} & \textbf{3398} & \textbf{1939} & \textbf{1576} & \textbf{1192} & \textbf{78} & \textbf{13448} & \textbf{100} \\
\hline
train4 & 4214 & 2719 & 1552 & 1262 & 955 & 63 & \textbf{10765} & 80 \\
dev4 & 261 & 168 & 95 & 76 & 58 & 2 & \textbf{660} & 5 \\
test4 & 790 & 511 & 292 & 238 & 179 & 13 & \textbf{2023} & 15 \\
\textbf{Total} & \textbf{5265} & \textbf{3398} & \textbf{1939} & \textbf{1576} & \textbf{1192} & \textbf{78} & \textbf{13448} & \textbf{100} \\
\hline
train5 & 4214 & 2720 & 1554 & 1263 & 956 & 64 & \textbf{10771} & 80 \\
dev5 & 261 & 168 & 95 & 76 & 58 & 2 & \textbf{660} & 5 \\
test5 & 790 & 510 & 290 & 237 & 178 & 12 & \textbf{2017} & 15 \\
\textbf{Total} & \textbf{5265} & \textbf{3398} & \textbf{1939} & \textbf{1576} & \textbf{1192} & \textbf{78} & \textbf{13448} & \textbf{100} \\
\hline
\cline{1-8}
\% & 39\% & 25\% & 14\% & 12\% & 9\% & 1\% & \textbf{100\%} \\
\cline{1-8}
\end{tabular}
\label{table:crossValidation}
\end{center}
\end{table}

\subsection{Preprocessing and Strategy Variations}

Preprocessing is carried out according to each shortening strategy. This study tested 10 strategies, as shown in Table \ref{table:strategies}. The first strategy is to take the first 70 tokens from the document. The second strategy is to take the 70 tokens in the middle of the document. The third strategy is to take the last 70 tokens from the document. The fourth strategy is to take the first 35 tokens and 35 tokens in the middle of the document. The fifth strategy is to take 35 tokens in the middle and the last 35 tokens of the document. The sixth strategy is to take the first 35 tokens and the last 35 tokens of the document. The seventh strategy is to take the first 30 tokens, 20 tokens of the middle, and the last 20 tokens of the document. The eighth strategy is to use human-annotated extractive summarization. The ninth strategy is to use human-annotated abstractive summarization. The tenth strategy is to use automatic abstractive summarization. In addition, this study tested the performance of full-text without shortening.

\begin{table}[htbp]
\caption{Shortening Strategies}
\begin{center}
\begin{tabular}{cl}
\hline
\textbf{No.} & \textbf{Strategy} \\
\hline
0 & No Shortening (Take All Tokens) \\
1 & Take First 70 Tokens \\
2 & Take Middle 70 Tokens \\
3 & Take Last 70 Tokens \\
4 & Take First 35 + Middle 35 Tokens \\
5 & Take Middle 35 + Last 35 Tokens \\
6 & Take First 35 + Last 35 Tokens \\
7 & Take First 30 + Middle 20 + Last 20 Tokens \\
8 & Extractive Summarization (23-220 tokens, Avg. 75 tokens) \\
9 & Abstractive Summarization (42-121 tokens, Avg. 68 tokens) \\
10 & Automatic Abstractive Summ. (43-88 tokens, Avg. 64 tokens)\\
\hline
\end{tabular}
\label{table:strategies}
\end{center}
\end{table}

\begin{figure}[htbp]
\centerline{\includegraphics[width=0.5\textwidth]{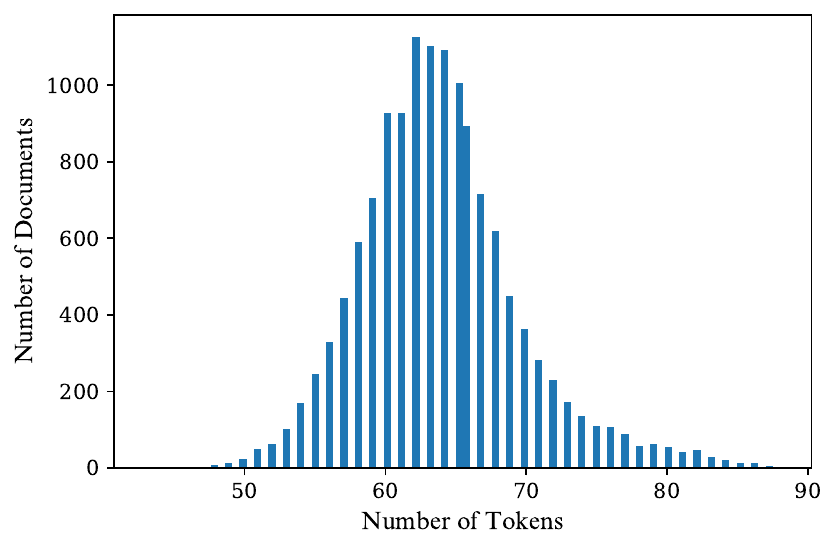}}
\caption{Histogram of Automatic Summarization}
\label{fig:automaticHistogram}
\end{figure}

To generate automatic abstractive summaries as the third variation of the summarization method, this study used a model provided by Wirawan \cite{Cahya2020Bert2BertSummarization}, a contributor of the repository of huggingface.co. The model is an encoder-decoder model based on 2 Indonesian monolingual pretrained BERT models. The contributor fine-tuned the model with the Liputan6 \cite{Koto2020Liputan6} dataset. The shortest summary generated has 43 tokens. The longest summary has 88 tokens. The average is 64 tokens. The distribution is as shown in Figure \ref{fig:automaticHistogram}. The parameters used to do summarization were as follows:
\begin{itemize}
    \item top\_k = 50
    \item top\_p = 0.95
    \item num\_beams = 10
    \item min\_length = 60
    \item max\_length = 120
    \item temperature = 0.8
    \item do\_sample = True
    \item use\_cache = True
    \item length\_penalty = 1.0
    \item early\_stopping = True
    \item repetition\_penalty = 2.5
    \item no\_repeat\_ngram\_size = 2
\end{itemize}

\subsection{Text Classification}

We fine-tune text classification models from a pretrained Indonesian DistilBERT model\footnote{\url{https://huggingface.co/cahya/distilbert-base-indonesian}}~\cite{Cahya2020DistilBERT}. DistilBERT itself~\cite{Sanh2019DistilBERT} is a distilled version of BERT~\cite{Devlin2019BERT}. We chose this base model because it is more efficient than the original BERT.

For the hyperparameter setting, we use 5 epochs, learning rate value 2e-5, batch size value 16, and the dropout probability value 0.1 \cite{Devlin2019BERT}. As for the optimizer algorithm and the loss function, we use AdamW and Categorical Cross Entropy respectively. The model is fine-tuned and evaluated with the train, validation, and test data from Filtered IndoSum which has been preprocessed according to each of the respective shortening strategies. The main metric used in this study is macro-averaged F1-score. If the F1-score evaluation at the end of each epoch is better than the previous one, the checkpoint is saved. This process is repeated for each of the five folds, and the F1-score is obtained by averaging the scores of all folds. We use Google Colab GPU (P100). To ensure reproducibility\footnote{Because Google Colab might assign different machines, setting a static random seed does not always help us obtain the same results every single time.}, we repeat the above experiment five times, and compute the final average F1-score.

\section{Result and Discussion} \label{sec:resultDiscussion}
This study finds that extractive summarization outperforms the majority of truncation variations in text classification tasks. There is only one truncation variation which is still better, that is taking the beginning of the document. This result is different from the result of Sun et al. \cite{Sun2019BertTextClassification} which concludes the head concatenated with the tail of the document. The main cause, in our opinion, is the difference in the types of documents or the length of truncation. The full result of this study is as shown in Table \ref{table:result}. F1-scores of the top 2 shortening strategies are close to F1-score of the full-text. The difference is less than 0.01.

\begin{table}[htbp]
\caption{Result}
\begin{center}
\begin{tabular}{clcc}
\hline
\textbf{Rank} & \textbf{Strategy} & \textbf{F1} & \textbf{Acc} \\
\hline
- & No Shortening (Take All) & 0.7863 & 92.35 \\
1 & Take First 70 & 0.7780 & 91.93 \\
2 & Extractive Summarization & 0.7773 & 91.33 \\
3 & Take First 35 + Last 35 & 0.7759 & 91.88 \\
4 & Take First 35 + Middle 35 & 0.7752 & 91.59 \\
5 & Abstractive Summarization & 0.7724 & 91.25 \\
6 & Take First 30 + Middle 20 + Last 20 & 0.7712 & 91.85 \\
7 & Automatic Abstractive Summariz. & 0.7667 & 91.03 \\
8 & Take Middle 35 + Last 35 & 0.7502 & 90.59 \\
9 & Take Middle 70 & 0.7424 & 89.73 \\
10 & Take Last 70 & 0.7370 & 89.86 \\
\hline
\end{tabular}
\label{table:result}
\end{center}
\end{table}

Table \ref{table:result} depicts that all truncation strategies involving the beginning of the document are ranked at the top. This indicates that the majority of news documents in our dataset contains the most important and/or relevant information at the beginning of the document. It also shows that taking the beginning of the document performs almost as well as when we take the document as a whole. On the other hand, the middle and the end of the document are equally less valuable. The abstractive summaries do not perform well, especially the automated ones.

As shown in Figure \ref{fig:extractiveHistogram}, \ref{fig:abstractiveHistogram}, and \ref{fig:automaticHistogram}, the extractive summarization strategy results in texts longer than the others, whereas the automatic abstractive summarization results is the shortest one, that is, in average less than the 70 tokens used by the various truncation strategies. Considering that the ``No Shortening'' strategy still yields the best result, we hyphotize that the shorter summaries degrades the overall performance, and that if we could control the length of the summarization system, it would have performed better.

The absence of dedicated GPU renders any statistical significance test requiring multiple runs of experiments infeasible in terms of time. However, we note that the use of cross-validation in our experiments reduces the variance of our results.

To take a deeper look, we match the tokens of the truncation strategy with the ground truth of extractive summarization, as shown in Table \ref{table:hit}. If the truncation has an intersection with extractive summarization of at least 1 token, it counts as a ``hit''. If it has no intersection at all it counts as ``miss''. Table \ref{table:hit} tells that the strategy ``Take First 70 Tokens'' has the least ``miss'' and is the best in Table \ref{table:result}. This shows that ``Take First 70 Tokens'' tends to be similar with extractive summarization and both are in the top 2.

\begin{table}[htbp]
\caption{Matching Truncated Documents with Extractive Summaries}
\begin{center}
\begin{tabular}{lrrc}
\hline
\textbf{Strategy} & \textbf{Hit} & \textbf{Miss} & \textbf{Total} \\
\hline
Take First 70 & 12988 & 460 & 13448 \\
Take First 35 + Middle 35 & 12791 & 657 & 13448 \\
Take First 30 + Middle 20 + Last 20 & 12595 & 853 & 13448 \\
Take First 35 + Last 35 & 12326 & 1122 & 13448 \\
Take Middle 70 & 4701 & 8747 & 13448 \\
Take Middle 35 + Last 35 & 3461 & 9987 & 13448 \\
Take Last 70 & 1401 & 12047 & 13448 \\
\hline
\end{tabular}
\label{table:hit}
\end{center}
\end{table}

We hypothesize that the IndoSum dataset benefits from the fact that, in the news domain, the first sentences are often the most important, resulting in the great performance of taking-the-first-70-tokens strategy. We believe that the extractive summarization method is potentially more generalizable in other types of dataset where the important parts often do not lie at the beginning of the document. However, due to the lack of availability of such a dataset, we leave the verification of this idea as a future work.

\section{Conclusion and Recommendation} \label{sec:conclusion}
This study concludes that extractive summaries as an alternative shortening strategy has a great potential for a Transformer-based classification model. However, it is not the best alternative for a uniform dataset like the Filtered IndoSum (Table \ref{table:result} and \ref{table:hit}), in the sense that most documents contain the main idea within their opening sentences. In this case, it is unsurprising that taking the first sentences performs best. In general, whenever the location of the main idea of the document is unknown, taking the first part of the document seems to be the best assumption. If one wants to bet by taking the middle or end of the document, we recommend making sure not to ignore the beginning of the document.

Nevertheless, in order to maximize the result of a classification task, the model cannot always rely on a general assumption of which part of the document to take. The potential of summarization needs to be exploited. In this study, automatic abstractive summarization has an unsatisfying performance. Future work needs to improve the performance of automatic extractive and abstractive summarization. On the other hand, a new dataset different to IndoSum, that is, with less uniformity of the main idea location, is necessary to further test the effectiveness of summarization as a shortening strategy in various kinds of long texts.

\bibliographystyle{./IEEEtran}
\bibliography{./conference_041818}

\end{document}